\documentclass[10pt, a4paper]{article}
\usepackage{lrec2022} % this is the new LREC2022 Style
\pdfoutput=1
\usepackage{multibib}
\newcites{languageresource}{Language Resources}
\usepackage{graphicx}
\usepackage{tabularx}
\usepackage{soul}
\usepackage{titlesec}
\titleformat{\section}{\normalfont\large\bfseries\center}{\thesection.}{1em}{}
\titleformat{\subsection}{\normalfont\SmallTitleFont\bfseries\raggedright}{\thesubsection.}{1em}{}
\titleformat{\subsubsection}{\normalfont\normalsize\bfseries\raggedright}{\thesubsubsection.}{1em}{}
\renewcommand\thesection{\arabic{section}}
\renewcommand\thesubsection{\thesection.\arabic{subsection}}
\renewcommand\thesubsubsection{\thesubsection.\arabic{subsubsection}}
\usepackage{times}
\usepackage{latexsym}
\usepackage{cmap} % поиск в PDF
\usepackage{mathtext} % русские буквы в формулах
\usepackage[T2A,T1]{fontenc} % кодировка
\usepackage[utf8]{inputenc} % кодировка исходного текста
\usepackage[russian,main=english]{babel} % локализация и переносы
\usepackage[table,xcdraw,dvipsnames]{xcolor}
\usepackage{microtype}
\usepackage{expex}
\usepackage{dblfloatfix}

\usepackage{caption}
\usepackage{enumitem}
\usepackage{amsmath}

\usepackage{mathptmx}

\DeclareRobustCommand{\cyrins}[1]{%
  \begingroup\fontfamily{erewhon-TLF}%
  \foreignlanguage{russian}{\textit{#1}}%
  \endgroup
}

\usepackage{epstopdf}
\usepackage[utf8]{inputenc}
\usepackage{todonotes}
\usepackage{hyperref}
\usepackage{xstring}
\usepackage{booktabs}
\usepackage{multirow}

\usepackage{color}

\title{Razmecheno: Named Entity Recognition from Digital Archive of Diaries  ``Prozhito''}

\newcommand{\hse}{\heartsuit}
\newcommand{\mipt}{\Diamond}
\newcommand{\airi}{\text{\textdaggerdbl}}
\newcommand{\iling}{\sharp}
\newcommand{\huawei}{\dagger}
\newcommand{\msu}{\spadesuit}
\newcommand{\sber}{\text{\textdollar}}

\makeatletter
\let\@fnsymbol\@arabic
\makeatother
\name{
Timofey Atnashev$^{\hse}$,\quad
Veronika Ganeeva$^{\hse}$,\quad
Roman Kazakov$^{\hse}$
\\
{\bf \large 
Daria Matyash$^{\hse\sber}$,\quad 
Michael Sonkin$^{\hse}$,\quad
Ekaterina Voloshina$^{\hse}$
} \\
{\bf \large 
Oleg Serikov$^{\hse\mipt\airi\iling}$,\quad
Ekaterina Artemova$^{\hse\huawei\msu}$
\medskip
}
}
\address{
$^{\hse}$ HSE University \quad
$^{\mipt}$ DeepPavlov lab, MIPT \quad
$^{\airi}$ AIRI \quad
$^{\iling}$ The Institute of Linguistics RAS \\
$^{\huawei}$ Huawei Noah’s Ark Lab \quad
$^{\msu}$ Lomonosov Moscow State University \quad
$^{\sber}$ Sber AI Centre\smallskip\\
\{taatnashev, vaganeeva, rmkazakov, dsmatyash, mvsonkin, eyuvoloshina\}@edu.hse.ru\\
\{oserikov, elartemova\}@hse.ru\smallskip\\
Moscow, Russia         
}

\abstract{
The vast majority of existing datasets for Named Entity Recognition (NER) are built primarily on news, research papers and Wikipedia with a few exceptions, created from historical and literary texts. What is more, English is the main source for data for further labelling. This paper aims to fill in multiple gaps by creating a novel dataset ``Razmecheno'', gathered from the diary texts of the project ``Prozhito'' in Russian. Our dataset is of interest for multiple research lines: literary studies of diary texts, transfer learning from other domains, low-resource or cross-lingual  named entity recognition. \\
Razmecheno comprises $1331$ sentences and $14119$ tokens, sampled from diaries, written during the Perestroika. The annotation schema consists of five commonly used entity tags: person, characteristics, location, organisation, and facility.  The labelling is carried out on the crowdsourcing platfrom Yandex.Toloka in two stages. First, workers selected sentences, which contain an entity of particular type. Second, they marked up entity spans. As a result $1113$ entities were obtained. Empirical evaluation of Razmecheno is carried out with off-the-shelf NER tools and by fine-tuning pre-trained contextualized encoders. We release the annotated dataset for open access.
 \\ \newline  
}

\begin{document}

\maketitleabstract

\section{Introduction} %KatyaA
Modern Named Entity Recognition (NER) systems are typically evaluated on datasets such as ACE, OntoNotes and CoNLL 2003, collected from news or Wikipedia. 
Other common setups to test NER systems include cross-lingual evaluation \cite{liang2020xglue} and evaluation in domains, other than  general, such as biomedical domain \cite{weber2020hunflair,wang2019cross}.

Additionally, the vast majority of NER dataset are in English. %misha: datasetS
A few large-scale datasets for other languages are NoSta-D \citelanguageresource{benikova2014nosta} (German), NorNE \citelanguageresource{jorgensen2020norne} (Norwegian), AQMAR \citelanguageresource{mohit2012recall} (Arabic), OntoNotes \citelanguageresource{hovy2006ontonotes} (Arabic, Chinese), FactRuEval \citelanguageresource{starostin2016factrueval} (Russian).

We present in this work a new annotated dataset for named entity recognition from diaries, written in Russian, -- ``Razmecheno''\footnote{``Got annotated''. The short form of the past participle neuter singular of the verb \cyrins{размечать} (``to annotate''). \url{https://github.com/hse-cl-masterskaya-prozhito/main}}. 
The texts are provided by the  project ``Prozhito''\footnote{``Got lived''. The short form of the past participle neuter singular of the verb \cyrins{прожить} (``to live''). \url{https://prozhito.org/}}, which digitizes and publishes personal diaries. 
Diaries exhibit different surface and style features, such as a complex narrative structure, an author-centricity, mostly expressed in simple sentences with predominance of verbs and noun phrases.

Design choices, made for the corpus construction, are the following. 
We follow the standard guidelines of named entity annotation and adopt four commonly-used types Person (\textsc{per}), Location (\textsc{loc}), Organization (\textsc{org}), Facility (\textsc{fac}).  We add one more type, \textsc{char}, which is used for personal characteristic (e.g. nationality, social group, occupation). Texts, used in the corpus, are sampled from the diaries, written in the late 1980's, the time period addressed as Perestroika.  We utilized crowd-sourcing to label texts. 

Our dataset enables assessing performance of the NER models in a new domain or in a cross-domain transferring. 
We make the following contributions: 
\begin{enumerate}
    \item We present a new dataset for Named Entity Recognition of 14119 tokens from 124 diaries from Prozhito. Entity types, used in the dataset, follow standard guidelines.  The dataset will be freely available for download under a Creative Commons ShareAlike 4.0 license at \url{https://github.com/hse-cl-masterskaya-prozhito/main};
    \item We assess the performance of the off-the-shelf NER taggers and fine-tuned BERT-based model on this data. 
\end{enumerate}

\section{Related work} % KatyaV
Most of the standard datasets for named entity recognition, as ACE \citelanguageresource{walker2005ace} and CoNLL \citelanguageresource{sang2003introduction}, consist of general domain news texts in English. For our study there are two related research lines: NER for the Russian language and NER in Digital Humanities domain.

\subsection{NER for Russian language}
\label{sec:nerus_review}
The largest dataset for Russian was introduced by \citelanguageresource{loukachevitch-etal-2021-nerel}. In NEREL, entities of types \textsc{per}, \textsc{org}, \textsc{loc}, \textsc{fac}, \textsc{gpe} (Geopolitocal entity), and \textsc{family} were annotated, and the total number of entities accounts to 56K. 

\citelanguageresource{starostin2016factrueval} presented FactRuEval for NER competition. The dataset included news and analytical texts, and the annotation was made manually for the following types: \textsc{per}, \textsc{org}  and \textsc{loc}. As of now, it is one of the largest datasets for NER in Russian as it includes 4907 sentences and 7630 entities.

Several other datasets for Russian NER, such as \textit{Named Entities 5}, WikiNER,  are included into project Corus\footnote{https://github.com/natasha/corus}.  Its annotation schema consists of 4 types: PER, LOC, Geolit (geopolitical entity), and MEDIA (source of information). Another golden dataset for Russian was collected by \citelanguageresource{gareev2013introducing}.  The dataset of 250 sentences was annotated for \textsc{per} and \textsc{org}.  For the BSNLP-2019 shared task, a manually annotated dataset of 450 sentences was introduced \citelanguageresource{piskorski2019second}.  The annotation includes \textsc{per}, \textsc{org}, \textsc{loc}, \textsc{pro} (products), and \textsc{evt} (events). 

Several silver datasets exist for Russian NER. WikiNEuRal \citelanguageresource{tedeschi-etal-2021-wikineural-combined} uses multilingual knowledge base and transfomer-based models to create an automatic annotation for \textsc{per}, \textsc{loc}, \textsc{prg}, and \textsc{misc}. It includes 123,000 sentences and 2,39 million tokens. In Natasha project, a silver annotation corpus for Russian Nerus\footnote{https://github.com/natasha/nerus} was introduced.  The corpus contains news articles and is annotated with three tags: \textsc{per}, \textsc{loc}, and \textsc{org}.  For Corus project, an automatical corpus WikiNER was created, based on Russian Wikipedia and methodology of WiNER \citelanguageresource{ghaddar2017winer}.

\subsection{NER applications to Digital Humanities}
% https://aclanthology.org/N19-1231.pdf
\citelanguageresource{bamman2019annotated} introduced LitBank, a dataset built on literary texts.  The annotation was based on ACE types of named entities and it includes the following types: \textsc{per}, \textsc{org}, \textsc{fac}, \textsc{loc}, \textsc{gpe} (geo-political entity) and \textsc{veh} (Vehicle).  The annotation was made by two of the authors for 100 texts. The experiments with models trained on ACE and on LitBank showed that NER models trained on the news-based datasets decrease significantly in the quality on literary texts.  \citelanguageresource{brooke2016bootstrapped} trained unsupervised system for named entity recognition on literary texts, which bootstraps a model from term clusters.  For evaluation they annotated 1000 examples from the corpus.  Compared to  NER systems, the model shows better results on the literary corpus data. 

Apart from English LitBank, a dataset for Chinese literary texts was created and described by \citelanguageresource{xu2017discourse}.  The dataset for Chinese literature texts had both rule-based annotation and machine auxiliary tagging, hence, only examples where gold labels and predicted labels differ were annotated manually.  The corpus of 726 articles were annotated by five people. Besides standard tags, as \textsc{per}, \textsc{loc}, and \textsc{org}, the authors used tags \textsc{thing}, \textsc{time}, \textsc{metric}, and \textsc{abstract}. 

Another approach to annotation was presented by \citelanguageresource{wohlgenannt2016extracting}.  The authors' purpose was to extract social networks of book characters from literary texts.  To prepare an evaluation dataset, the authors used paid micro-task crowd-sourcing. The crowd-sourcing showed high quality results and appeared to be a suitable method for digital humanities tasks.

\section{Dataset collection} 
\subsection{Annotation schema}
Our tag set consists of five types of entities. 
%The choice of classes is motivated by their semantics (taxonomic class) and distribution in contexts.
This tag set was designed empirically for texts of diaries from common tags used in related works \citelanguageresource{walker2005ace,bamman2019annotated}.
\begin{itemize}
    \item \textsc{\textbf{per}}: names/surnames of people, famous people and characters are included (see Example \ref{ex:per});
    \item \textsc{\textbf{char}}: characteristics of people, such as titles, ranks, professions, nationalities, belonging to the social group (see Example \ref{ex:char_fac});
    \item \textsc{\textbf{loc}}: locations/places, this tag includes geographical and geopolitical objects such as countries, cities, states, districts, rivers, seas, mountains, islands, roads etc. (see Example \ref{ex:loc});
    \item \textsc{\textbf{org}}: official organizations, companies, associations, etc. (see Example \ref{ex:org});
    \item \textsc{\textbf{fac}}: facilities that were built by people, such as schools, museums, airports, etc. (see Example \ref{ex:char_fac});
    \item \textsc{\textbf{misc}}: other miscellaneous named entities.
\end{itemize}

These five entity types can be clearly divided into two groups: the first one, \textsc{per}-\textsc{char}, is related to people and the second one,  \textsc{org}-\textsc{loc}-\textsc{fac}, is related to places and institutions.  

We annotated flat entities, so that the overlap between two entities is not possible. The main principle of the annotation is to choose as long as possible text span for each entity, not to divide them when not required, because our schema does not assume multi-level annotation, when one entity can include another ones. For example, a name and a surname coalesce in one PER entity, rather than being two different ones (see Example \ref{ex:per}). % пример

%А ведь Леон просил меня отозваться лишь о Жаке Ланге .

\ex
\label{ex:per}
\begingl
    \gla \foreignlanguage{russian}{А}
    \foreignlanguage{russian}{ведь} \foreignlanguage{russian}{Леон} \foreignlanguage{russian}{просил} \foreignlanguage{russian}{меня} \foreignlanguage{russian}{отозваться}
    \foreignlanguage{russian}{лишь}
    \foreignlanguage{russian}{о}
    \foreignlanguage{russian}{Жаке Ланге}//
    \glb And really $\underbrace{\textbf{Leon}}_{\text{\textsc{per}}}$ asked me to.talk only about $\underbrace{\textbf{Jack Lang}}_{\text{\textsc{per}}}$//
    \glft `And Leon asked me to talk only about Jack Lang'.//
\endgl
\xe

\ex
\label{ex:loc}
\begingl
    \gla \foreignlanguage{russian}{Орёл}
    \foreignlanguage{russian}{самый} \foreignlanguage{russian}{литературный} \foreignlanguage{russian}{город} \foreignlanguage{russian}{в} \foreignlanguage{russian}{России} //
    \glb $\underbrace{\textbf{Orel}}_{\text{\textsc{loc}}}$ the.most literary city in $\underbrace{\textbf{Russia}}_{\text{LOC}}$//
    \glft `Orel is the most literary city in Russia'.//
\endgl
\xe

\ex
\label{ex:org}
\begingl
    \gla \foreignlanguage{russian}{Позвонил} \foreignlanguage{russian}{в} \foreignlanguage{russian}{``Урал'':} \foreignlanguage{russian}{надо} \foreignlanguage{russian}{все-таки} \foreignlanguage{russian}{дать} \foreignlanguage{russian}{им} \foreignlanguage{russian}{знать} \foreignlanguage{russian}{о} \foreignlanguage{russian}{моем} \foreignlanguage{russian}{прилете.} //
    \glb called in $\underbrace{\textbf{``Ural''}}_{\text{\textsc{org}}}$ need after.all give them know about my arrival //
    \glft `I called the ``Ural'': after all I have to let them know about my arrival'.//
\endgl
\xe

\ex
\label{ex:char_fac}
\begingl
    \gla \foreignlanguage{russian}{Солдаты} \foreignlanguage{russian}{живут} \foreignlanguage{russian}{в} \foreignlanguage{russian}{вагоне} \foreignlanguage{russian}{на} \foreignlanguage{russian}{этой} \foreignlanguage{russian}{станции}. //
    \glb $\underbrace{\textbf{soldiers}}_{\text{\textsc{char}}}$ live in car on this $\underbrace{\textbf{station}}_{\text{\textsc{fac}}}$//
    \glft `Soldiers live in a car at this station'.//
\endgl
\xe

\begin{figure*}[!ht]
	\centering
	\includegraphics[width=\textwidth]{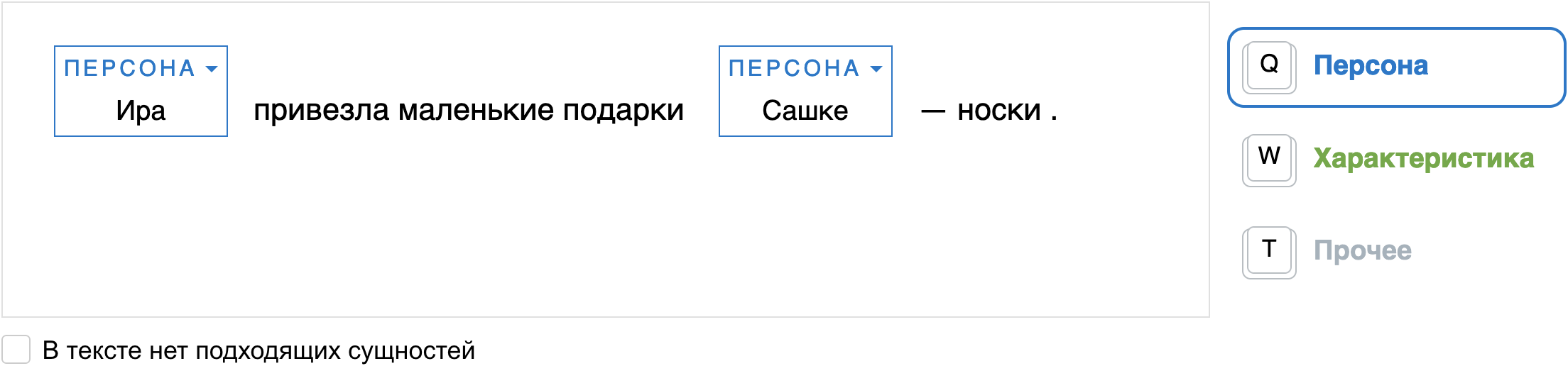}
	\caption{Annotation of a phrase given in Yandex.Toloka: Ira brought socks as small presents for Sasha.\\ Available annotations are: Person (blue), Characteristics (green), Misc (grey), no entities present (checkbox)}
	\label{fig:symspell_time}
\end{figure*}

In ambiguous cases entity tags were identified based on the context, so the same entity in different sentences could be tagged as two different types, for instance, \textit{university} could be annotated as \textsc{org} or \textsc{fac} . If an entity was used in a metaphorical sense, it would not be annotated with any tag.

\ex
\label{fourth}
\begingl
    \gla \foreignlanguage{russian}{Будет} \foreignlanguage{russian}{и} \foreignlanguage{russian}{на} \foreignlanguage{russian}{нашей} \foreignlanguage{russian}{улице} \foreignlanguage{russian}{праздник} //
    \glb will and on our street a.festival//
    \glft `Every dog has its own day'.//
\endgl
\xe

\subsection{Preliminary markup}

% In order to get the NER markup, it would be possible to give all the texts for markup to assessors, but obviously this does not make much sense, since it is long, expensive and many sentences would be marked up in which there are no entities at all. 
% Therefore, it was decided to train an automatic binary classifier that determines whether there are entities in a sentence. 

% At first, a small amount of data was collected, from which it became clear that in many sentences there are no entities at all. In order to avoid this and not to give such sentences to assessors for markup, it was decided to train a binary classifier (on which sample???) that determines the presence of entities in the sentence.
We performed preliminary analysis of the random subsets of the ``Prozhito'' corpus. The analysis revealed that most of the sentences contain no entities at all.
To avoid costly looping over all sentences, we developed a two-stage annotation pipeline. The first stage aims at selecting sentence candidates, which may include entities of interest. This helps to reduce the amount of sentences, sent to crowd workers and exclude sentences with no entities at all. During the second stage entity spans are labeled in the pre-selected candidates from the first stage.

Two classifiers were trained on a small manually annotated training set --- for \textsc{per}-\textsc{char}  and \textsc{org}-\textsc{loc}-\textsc{fac}  groups, respectively. The task of these classifiers is to predict, whether an entity from a group is present in a sentence, or not. These classifiers do not aim at entity recognition, but rather at binary entity detection.
% We considered two subsets of tags, where one is related to people (PER, CHAR) and the second one is related to different locations (LOC, ORG, FAC). 

% However, due to the differences between two entities groups - LOC, ORG, FAC and PER, CHAR, were implemented classifiers for each of this group, respectively. 
% \oleg{this seems like a wrong place to introduce the alternation of PERCHAR vs ORGLOCFAC. where should we put it instead?}

We leverage upon four possible base models as classifiers:  \texttt{ruBERT-tiny\footnote{\url{https://huggingface.co/cointegrated/rubert-tiny}}}, \texttt{ruBERT\footnote{\url{https://huggingface.co/DeepPavlov/rubert-base-cased}}}, \texttt{ruRoBERTa\footnote{\url{https://huggingface.co/sberbank-ai/ruRoberta-large}}}, \texttt{XLM-RoBERTa\footnote{\url{https://huggingface.co/xlm-roberta-base}}}. Table  \ref{tab:binary_scores} presents with the classification scores. 
A small number of marked up sentences (198) were taken as test sample.

% \begin{figure}[htb]
%     \centering
%     \includegraphics[width=0.45\textwidth]{img/roberta_scores_table.png}
%     \caption{Caption\todo{migrate from figure to table,  recalculate numbers w frozen train-test-val data}}
%     \label{tab:binary_models_scores}
% \end{figure}

\begin{table}[htb]
\resizebox{0.45\textwidth}{!}{

\begin{tabular}{lccc}
\toprule
                     \textbf{Models}& \textbf{Precision} & \textbf{Recall} & \textbf{Micro f1-score} \\\midrule
\texttt{ruBERT-tiny} & 0.81      & 0.88   & 0.84     \\
\texttt{ruBERT}      & 0.89      & 0.91   & \textbf{0.90}     \\
\texttt{ruRoBERTa}   & \textbf{0.90}      & 0.88   & 0.89     \\
\texttt{XLM-RoBERTa} & 0.80      & \textbf{0.99}   & 0.89    \\

\bottomrule
\end{tabular}
}
\caption{Transformer-based binary classifiers scores}
\label{tab:binary_scores}
\end{table}

% It is worth to pay attention to the fact that the additional training of ruBERT did not give significant improvements, since after it the confidence of the model on examples with a prediction probability of $> 0.9$ in some cases fell to $0.6$, which indicates a fairly large variety of sentences in sample as well as as in the corpus itself. 

As a result, \texttt{ruRoBERTa} was chosen as the base model. In this task, the precision is more important than the recall, since we markup only part of the corpus and, therefore, we still miss some of the information, but at the same time we want to have any entities in the selected sentences with a high probability. It is worth mentioning that this approach allowed to save big amount of money, because the most optimal solution for maximizing the recall would be to give all the data to assessors for markup. 
% \timofey{\foreignlanguage{russian}{про деньги}}

To train both classifiers, a random sample of size 1500 was taken from diaries belonging to Perestroika period. Texts were independently marked up by assessors for the presence of \textsc{org}-\textsc{loc}-\textsc{fac}  and \textsc{per}-\textsc{char}. Due to the fact that it was important to achieve a balance of classes in the training sample, and there were more texts with \textsc{per}-\textsc{char} then \textsc{org}-\textsc{loc}-\textsc{fac} , the training samples for \textsc{org}-\textsc{loc}-\textsc{fac}  and \textsc{per}-\textsc{char} turned out to be different - 829 and 1465 records accordingly (see Table \ref{tab:roberta_binary_detailed} for the validation set scores).

% On the validation set following results were achieved: 
All available sentences were marked up by binary classifier and after that were chosen sentences with following conditions: 
\begin{enumerate}
    \item In sentence there are entities from \textsc{per}-\textsc{char} and \textsc{org}-\textsc{loc}-\textsc{fac} groups, respectively; 
    \item Classifier was the most confident on these sentences.
\end{enumerate}
% \begin{figure}[htb]
%     \centering
%     \includegraphics[width=0.2\textwidth]{img/oleg_timofey_orglocfac.png}
%     \includegraphics[width=0.2\textwidth]{img/oleg_timofey_perchar.png}
%     \caption{Caption\todo{migrate to table, recalculate numbers w frozen train-test-val data}}
%     \label{fig:my_label}
% \end{figure}

\begin{table}[!ht]
\begin{tabular}{lccc}
\toprule
            \textbf{Entity Type}& \textbf{Precision} & \textbf{Recall} & \textbf{F1-score} \\\midrule
\textsc{org}-\textsc{loc}-\textsc{fac}  & 0.94      & 0.92   & 0.94     \\
\textsc{per}-\textsc{char}   & 0.89      & 0.81   & 0.82     \\
\bottomrule
\end{tabular}
\centering
\caption{\texttt{ruRoBERTa} scores in the binary classification task}
\label{tab:roberta_binary_detailed}
\end{table}

Most confidence here means the average probabilities of the each entities groups.  Finally, the sentences selected this way were given to the assessors for further marking.

\subsection{Crowd-sourcing annotation}
\paragraph{Annotation setup}
For annotation we used Russian crowd-sourcing platform  Yandex.Toloka \footnote{\url{https://toloka.yandex.ru/}}. 
We prepared two tasks for assessors: determination of \textsc{per}-\textsc{char} and of \textsc{org}-\textsc{loc}-\textsc{fac} in ``Prozhito'' texts. 
The task required native knowledge of the Russian language for satisfactory solution.
Before annotation it is necessary to get through the learning pool with hints (20 sentences) and an exam (10 sentences) that show whether assessors understand the meaning of the given NE tags. 
The sentences were tokenized with Razdel tokenizer\footnote{\url{https://github.com/natasha/razdel}}.

The tasks for learning, exam and control were initially annotated by the co-authors with help of annotation tool BRAT\footnote{\url{https://brat.nlplab.org/}}.

% %
% \begin{figure*}[th!]
%   \centering
%   \includegraphics[width=\textwidth]{img/toloka_task_example_1.png}
%   \caption{Number of RSUs that each vehicle has encountered}
%   \label{fig:RSUencountered}
% \end{figure*}

% \noindent%
% \begin{figure*}
% \begin{minipage}{\textwidth}% to keep image and caption on one page
% \makebox[\textwidth]{%        to center the image
%   \includegraphics[width=\textwidth]{img/toloka_task_example_1.png}}
% \captionof{figure}{Annotation of a phrase given in Yandex.Toloka}\label{fig:symspell_time}%      only if needed  
% \end{minipage}
% \end{figure*}

If an annotator succeeded in learning and exam (mark $\geq$ 50\% for learning and $\geq$ 80\% for exam), he/she got a special skill which allowed to start assessment of sentences in the main pool. 
Our main pools in both tasks consist of approximately $1500$ tasks and $400$ control sentences. Tasks were given to annotators on pages, Figure \ref{fig:symspell_time} depicts the task interface. Each page consisted of $4$ normal tasks and $1$ control task. A fee for one page was 0.05\$. The average time of completion of page was about one minute. Overall, the fee per hour exceeded minimum wage in Russia.
The overlap for each sentence given in Toloka is 3 in order to choose the most popular variant of markup as a correct one.
Control tasks are necessary for monitoring of an annotation quality.
We banned users if they skipped more than $7$ task suites in a row or if they had less than $30\%$ correct control responses.

% можно в аппендиксе в секции чтомыдсеаллаинетак рассказать e.g. о том, что мы вот использовали 2 среды , а лучше бы сразу ту, в которой работали краудсорсеры
% todo каким-то образом представить согласие разметчиков (text: почему мы не использовали стандартную аггрегацию; 
% viz: (варианты: все согласны, все несогласны тогда росло перекрытие, 1:2 тогда выбирали мнение большинства) 1:2 можно визуализировать: оси -- теги, клетки -- доля случаев спора между ними)

% я помню, что мы говорили про скриншот(-ы) с Толоки, но теперь я не понимаю, реально ли они нам нужны и какие взять, если всё-таки нужны

% я решила показать пример, как можно в целом добавлять картинки/иллюстрации с подписью. ну и вообще добавила, что посчитала нужным. в  includegraphics нужно прописать название изображения Д.М.

\paragraph{Annotators agreement analysis}

% 1. процентное соотношение -- сколько примеров разметились невсесогласием
% OLF
% 1    0.627119
% 2    0.309927
% 3    0.062954

% PC
% 1    0.636494
% 2    0.277299
% 3    0.070402
% 4    0.012931
% 6    0.001437
% 5    0.001437

While in most of the cases annotators had no dispute, voting mechanism has been involved in nearly one third of cases provided in the corpus ($38\%$ in the \textsc{org}-\textsc{loc}-\textsc{fac} task, $36\%$ in \textsc{per}-\textsc{char} tasks, respectively).

In both tasks typical annotators disagreement pattern was two competing annotation hypotheses. 
In the \textsc{org}-\textsc{loc}-\textsc{fac} task, that was mostly caused by different labels plausible for certain rare events. The ability to correctly disambiguate such terms relied onto rather rare factual knowledge, thus provoquing annotation errors (as in \cyrins{Сижу в гостинице ``Одесса''.} (`Staying in the hotel ``Odessa'''.), the challenging choice is `hotel ``Odessa''' is a \textsc{fac} or an \textsc{org} entity).
While the same group of annonators disagreements was found in the \textsc{per}-\textsc{char} task, there also emerged two more disagreements patterns: (i) identifying the proper span for the characteristics (annotating the whole \cyrins{полковник в отставке} (`the retired colonel') or only \cyrins{полковник} (`colonel') ) and (ii) inaccurate boundaries detection for persons initials, which mostly emerged when the annontators missed to higlight the dot in the name shortenings (as with \cyrins{М . С .} in \cyrins{М . С . его очень ценил поначалу.} (`M.S. valued him a lot in the beginning') ) . 

\begin{table}[!b]

    \centering
    \resizebox{\linewidth}{!}{
    \begin{tabular}{lcccc}
    \toprule
    \textbf{Type}& \textbf{\# Entities} & \textbf{\% Entities} & \textbf{\# Mentions} & \textbf{\% Mentions} \\
    \midrule
    \textsc{char} & 282 & 25.0\% & 290 & 19.7\% \\
    \textsc{fac} & 71 & 6.4\% & 106 & 7.2\% \\
    \textsc{loc} & 186 & 16.7\% & 221 & 15.0\% \\
    \textsc{org} & 73 & 6.6\% & 137 & 9.3\% \\
    \textsc{per} & 490 & 44.0\% & 708 & 48.0\% \\
    \textsc{misc} & 11 & 1.0\% & 12 & 0.8\% \\
    \midrule
    \textbf{Total} & \textbf{1113} & \textbf{100.0\%} & \textbf{1474} & \textbf{100.0\%} \\
    \bottomrule
    \end{tabular}
    }
    \caption{Dataset entities statistics}

    \label{tab:entities_stats}
\end{table}
Rare cases with more than two competing annotations were mostly of random nature (as with birds being annotated as \textsc{per}), or caused by the appearance of rare words (as with calzones being annotated as Person).

\subsection{Dataset statistics}

\begin{table*}[t]
    \centering
    %\resizebox{0.45\textwidth}{!}{
    \resizebox{\textwidth}{!}{
    \begin{tabular}{lp{15cm}}
    \toprule
    \textbf{Entity Type} & \textbf{Top-10 mentions}\\
    \midrule
    \textsc{char} &  \cyrins{ребёнок} (`a child'), \cyrins{женщина} (`a woman'), \cyrins{президент} (`a president'), \cyrins{друг} (`a friend'), \cyrins{поэт} (`a poet'), \cyrins{папа} (`a dad'), 
    \cyrins{писатель} (`a writer'), \cyrins{жена} (`a wife'), \cyrins{отец} (`a father'), \cyrins{военный} (`a military')\\
    \midrule
    \textsc{fac} & \cyrins{театр} (`a theatre'), \cyrins{аэропорт} (`an airport'), \cyrins{дом} (`a house'), \cyrins{школа} (`a school'), \cyrins{музей} (`a museum'), \cyrins{кафе} (`a cafe'), \cyrins{станция} (`a station'), \cyrins{библиотека} (`a library'), \cyrins{посольство} (`an embassy'), \cyrins{тюрьма} (`a prison') \\
    \midrule
    \textsc{loc} & \cyrins{город} (`a city'), \cyrins{Москва} (`Moscow'), \cyrins{Россия} (`Russia'), \cyrins{улица} (`a street'), \cyrins{Ленинград} (`Leningrad'), 
    \cyrins{проспект} (`an avenue'), \cyrins{Кандагар} (`Kandagar'), \cyrins{озеро} (`a lake'), \cyrins{страна} (`a country'), \cyrins{запад} (`west') \\
    \midrule
    \textsc{org} &\cyrins{ЦК} (`Central Committee'), \cyrins{совет} (`a council'), \cyrins{парламент} (`a parliament'), \cyrins{Политбюро} (`Politburo'), \cyrins{Правда} (`Pravda'), \cyrins{КПСС} (`the Communist Party of the Soviet Union'), \cyrins{издательство} (`a publishing house'), \cyrins{верховный} (`supreme'), \cyrins{Мосфильм} (`Mosfilm'), \cyrins{союз} (`a union') \\
    \midrule
    \textsc{per} & \cyrins{Горбачев} (`Gorbachev'), \cyrins{Борис} (`Boris'), \cyrins{Ельцин} (`Yeltsin'), \cyrins{Володя} (`Volodya'), \cyrins{Таня} (` Tanya'), \cyrins{Витя} (`Vitya'), \cyrins{Рыжков} (`Ryzhkov'), \cyrins{Яковлев} (`Yakovlev'), \cyrins{Сергей} (`Sergey'), \cyrins{Иван} (`Ivan') \\
    \bottomrule
    \end{tabular}
    }
    %}
    \caption{Top-10 mentions for each entity type}
    \label{tab:top10}
\end{table*}
% в ноушене записано, какие мы статистики хотим вклчюить 
% вот оно из ноушена
% гистограмма какой тег сколько раз встретился, + топ-10 строк для каждого тега
% у тегов в гистограмме отрезаем {BI}- префикс
% длины предложений — распределение
% референсные статьи
% - норвежская — согласие между аннотаторами (хитмапы про то, как аннотаторы путали теги); ванильная статистика типа длин

\begin{figure}[!ht]
    \includegraphics[height=6cm]{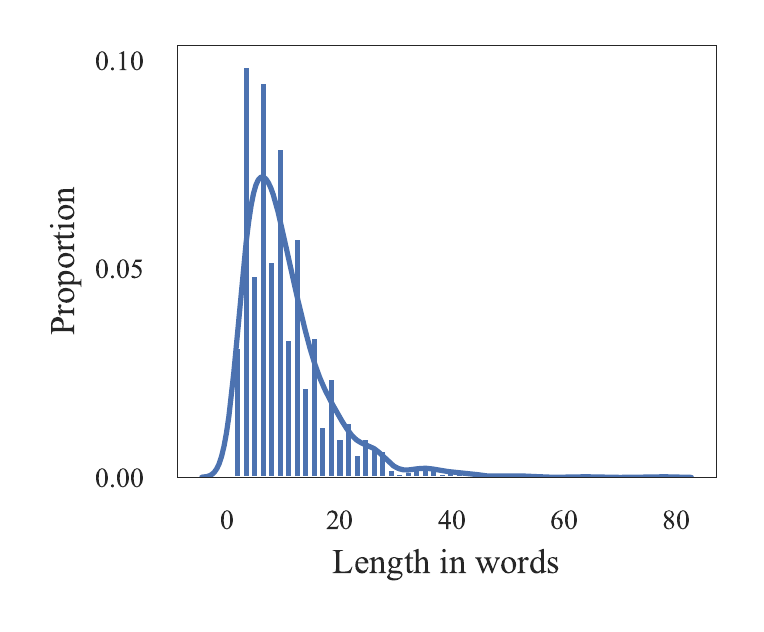}
    \caption{Distribution of tags}
    \centering
    \label{fig:lengths}
\end{figure}

The total number of sentences in the dataset is $1331$ and the total number of tokens is $14119$. The average sentence length is $10.61$ tokens (see Figure \ref{fig:lengths}).  $1113$ entities were identified at all ($1474$ mentions). The average length of entity in tokens is $1.32$ token.
% если с пункиуацией то 17508

\begin{figure}[!ht]
\includegraphics[height=6cm]{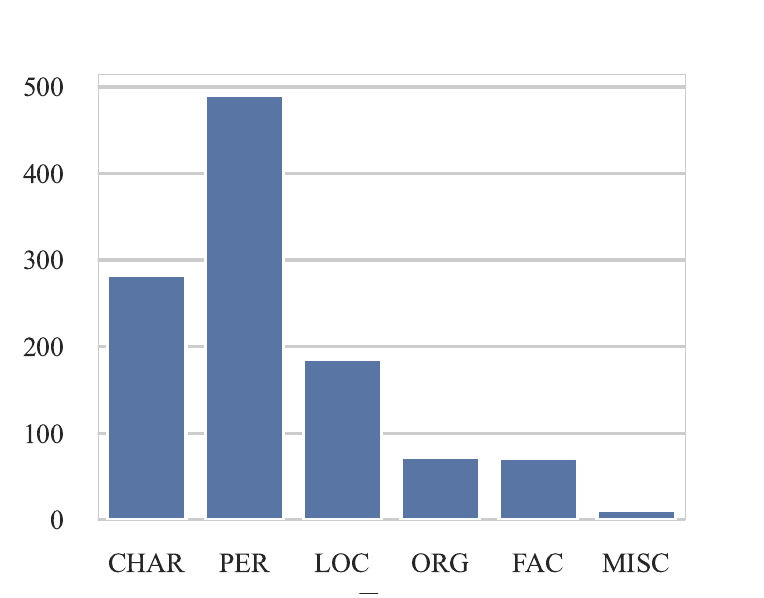}
\caption{Distribution of entity types}
\centering
\label{fig:tags}
\end{figure}

Table \ref{tab:entities_stats} and Figure \ref{fig:tags} describe dataset statistics. \textsc{per} is the most frequent tag, a little less than a half of all entities are of this type. Persons are often provided via a few tokens. The rest of types do not represent the same variance between mentions and entities. \textsc{MISC} entities are only 1\% of all entities.

\footnote{Notably, PER class contains lots of multiple words sentences}

Popular mentions of entities actually represent concepts and personalities of Perestroika period (see Table \ref{tab:top10}). As we can see, there are main politic figures in the list (e.g. Boris Yeltsin, Mikhail Gorbachev, Nikolai Ryzhkov) as well as old soviet political authorities (e.g. Central Committee, the Communist Party of the Soviet Union, Politburo). Some words that were new at that time, such as `a president' (since Gorbachev became the first president of USSR in 1990) are among the most frequent words. Another important trend is the discussion of the Soviet-Afghan war, as Kardagan was one of the centres of soviet troops' dislocation.

\section{Evaluation}  % Nika + Dasha

% todo написать про берт (Тимофей и Даша); дописать про off-the-shel (катя В);
% (Катя, Тимофей, Даша) план тетради с трансформерами 3 тетради(1. предобработка (тут ещё толока->BIO) 2. инференс 3. анализ результатов)
% структура типичной тетради: сначала инсталлы. потом импорты, потом константы с путями (дальше никакго хардкода путей). потом функции, которые будут запущены ниже (можно структурировать по смыслу)

% готовые и0ты поддерж те же самые теги, давайте посмотрим вдруг они норм работают (посмотрим на конф матрицу, посмотрим на скоры на тесте (есть ли у нас стандартный зафиксированный трейн-тест сплит) )

% картинки могут быть совместные чтобы было удобнее сравнивать

% \selectlanguage{russian}
% \begin{itemize}
%     \item стандартный train-test-val split датасета
%     \item матрица ошибок
%     \item микро ф1- мера на test данных
%     \item итоги. библиотека z победила
%     % двавате теперь обучим берт
%     \item обучить берта
%     \item матрицы ошибок, ф-меры
%     \item разные кодировщики
%     \begin{itemize}
%         % список моделей которые уже запущены
%         \item tinybert-ru
%         \item rubert
%         \item ruroberta large
%         % туду
%         \item xlm-r base
%         % ruBERT, roberta, xlmr base обсчитать в бейз-версиях
%     \end{itemize}
%     \item вывод
% \end{itemize}
% \selectlanguage{english}

\subsection{Off-the-shelf tools}
We use a selection of of publicly available,  NER systems: \texttt{Natasha-Slovnet}, \texttt{Stanza}, \texttt{SpaCy}. 

\texttt{Slovnet} is a neural network based tool for NLP tasks, including NER annotation. 
\texttt{Slovnet} is a part of Natasha project. \footnote{\url{https://github.com/natasha/slovnet\#ner}} 
\texttt{Slovnet}'s annotation includes \textsc{per}, \textsc{loc} and \textsc{org}.

\texttt{Stanza} is a Stanford state-of-art model \footnote{\url{https://stanfordnlp.github.io/stanza/}}. 
\texttt{Stanza} is based on Bi-LSTM model and CRF-decoder. 
\texttt{Stanza} for Russian is a 4-entity system, which includes \textsc{per}, \textsc{loc}, \textsc{org} and \textsc{misc}.

NER system developed by \texttt{SpaCy} is a transition-based named entity recognition component. 
We use \texttt{natasha-SpaCy} \footnote{\url{https://github.com/natasha/natasha-spacy}} model trained on two resources - Nerus \footnote{\url{https://github.com/natasha/nerus}} and Navec \footnote{\url{https://github.com/natasha/navec}}. 
\texttt{Natasha-SpaCy} model can detect \textsc{per}, \textsc{loc} and \textsc{org} entities in our dataset. % \footnote{\url{https://github.com/natasha/natasha-spacy}} ORG LOC PER

We have compared results of these models on our dataset.
\begin{table}[!ht]
    \centering
    %\resizebox{\linewidth}{!}
    \begin{tabular}{lcccc}
    \toprule
    \textbf{Models}  & \textbf{\textsc{per}}  & \textbf{\textsc{loc}}  & \textbf{\textsc{org}} & \textbf{Overall}  \\
    \midrule
    \texttt{SpaCy}   & 0.64          & \textbf{0.54} & \textbf{0.16} & 0.95\\
    % \midrule
    \texttt{Stanza}  & 0.69          & 0.4           & 0.11  & 0.94   \\
    % \midrule
    \texttt{Natasha} & \textbf{0.77} & \textbf{0.54} & 0.14 & \textbf{0.96} \\    
    \bottomrule
    \end{tabular}
    %}
    \caption{The performance of off-the-shelf tools}
    \label{tab:off-the-shelf}
\end{table}

As seen from the table \ref{tab:off-the-shelf}, \texttt{Natasha-Slovnet} showed the best perfomance on our dataset for \textsc{per} and \textsc{loc}, while SpaCy was the best on \textsc{loc} and \textsc{org} detection. 
However, the results of all models are significantly worse than the results on other datasets (Appendix \ref{app:comparison}). 
Such results prove our hypothesis that off-the-shelf tools do not recognise entities on a diary-based dataset, for they were trained on news data.
\begin{figure*}[!ht]
\centering
\includegraphics{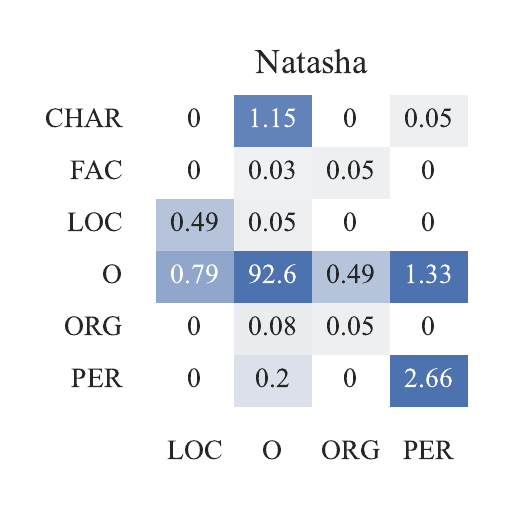}
\includegraphics{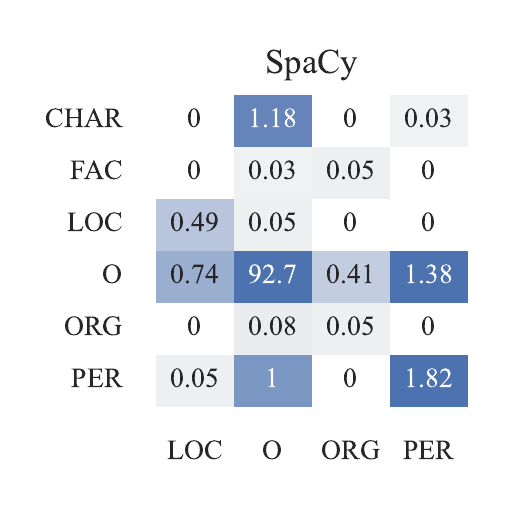}
\includegraphics{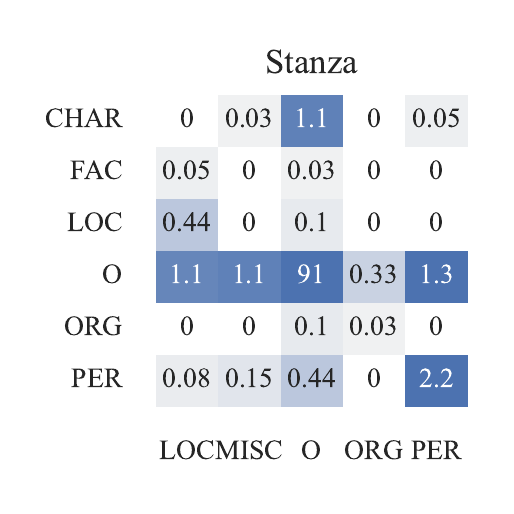}
\caption{Confusion matrix for off-the-shelf tools per token in relative weights}
\label{fig:image}
\end{figure*}

A closer look at the models' performance (figure \ref{fig:image}) reveals what caused  entity recognition issues.
Most of the models often detect false \textsc{loc} and \textsc{per} entities. 
In this case \texttt{SpaCy} shows the best results. 
\texttt{Natasha-Slovnet} has the greatest recall especially on \textsc{loc} and \textsc{per}. 
All models often annotated \textsc{org} as a non-entity. 
As our texts come from diaries written in 1990s, some organisations could not exist anymore, and models do not recognise them.  

\textsc{fac} and \textsc{char} were not on the entity lists of the models, therefore, the models did not recognise these tags. However, we would expect the models to mark \textsc{char} as \textsc{per} and \textsc{fac} as \textsc{loc} or \textsc{org} because those tags are related.
Indeed this happens for \textsc{fac} but not for \textsc{char}.
This happens as most of named entities are proper nouns and start with capitalized letters, unlike \textsc{char}.
All models annotated \textsc{fac} more often as \textsc{org} than as non entity. 

Another problem is caused by false detection of named entities' span boundaries. 
To account for this, we introduced the following approach. 
We counted all cases when models did not found entities at all, detected false entities or used a wrong tag (combined as `false detected') or models included more or less words from one or both sides. 
\texttt{Natasha} showed the best results, for it detects right boundaries for the most of the spans. 
The most common error though for all models was not finding an entity. 
Other mistakes include a shift of boundaries to the left and including more or less words on the left side especially for \textsc{per} recognition. 
It could be possibly explained that textsc{char} entity proceeds \textsc{per} entity (for instance, 
% \foreignlanguage{russian}{профессор Иванов} 
\cyrins{\textit{профессор Иванов}} (`professor Ivanov') where `professor' is \textsc{char}). 
Off-the-shelf models do not include textsc{char} entity and could annotate them as \textsc{per}. 
Problems of narrower boundaries could be caused by excluding quoting markers in automatic annotation. 

\subsection{Fine-tuned models}

% даша и тимофей
We fine-tuned multiple Transformer models for NER:  \texttt{ruBERT}, \texttt{ruBERT-tiny}, \texttt{ruRoBERTa}, \texttt{XLM-RoBERTa}. The performance was evaluated according to F1-scores per named entity and overall micro F1-score.

We used weighted cross-entropy as a loss function. Inverse tag frequency was taken as weights for cross-entropy, which helped us gain better better results on unbalanced data. 
We also sorted the dataset by the length of tokens and than split it to batches, which slightly improved models' performance. Models were trained in an unfrozen manner.
The detailed hyperparameters values used to train the models are provided in the Appendix~\ref{app:trans_hyperparam}. 
The performance  was  evaluated according to per-class and overall micro-averaged F1-score.

% Number of epochs, learning rate and weight decay for model training were chosen by the supervized Grid Search strategy. 
% We also tried different approaches to unfreeze layers, all the shown fine-tuned models used all freezed pre-trained layers except the last one.

%Additionally, we fine-tune several transformer models (namely, ruBERT, ruBERT-tiny, ruRoBERTa, XLMR) on the collected dataset.

\subsection{Results}

\texttt{Natasha} had the best F1-score among all off-the-shelf tools. Nevertheless, results achieved for our corpus are below \texttt{Natasha}'s results on news-based datasets. 

Fine-tuned transformers showed better results than off-the-shelf tools. Predictions made by \texttt{XLM-RoBERTa} had the highest overall F1-score, \texttt{ruBERT} model's performance had the best F1-scores for most tags (\textsc{fac}, \textsc{loc}, \textsc{org}) and top-3 best results for \textsc{char} and \textsc{per} tags. 
According to Table \ref{tab:transformers_comparison_large}, we can consider \texttt{ruBERT} the best model for our datasets as it successfully predicts major and minor classes. 
%  \dasha{\foreignlanguage{russian}{ (done)надо понятнее объяснить, почему у разных моделей разное колво эпох (эрли стоппинг?) + сослаться где-то в тексте на таблицу}}.

% \begin{table}[]
%     \centering
%     \resizebox{\linewidth}{!}{
%     \begin{tabular}{lcc}
%         \toprule
%         \textbf{Models} & \textbf{Number of epochs} & \textbf{Overall F1-score} \\
%          \midrule
%          \textbf{ruBERT-tiny} & 50 & 0.731 \\
%          \textbf{ruBERT} & 10 & \textbf{0.813} \\
%          \textbf{ruRoBERTa} & 5 & 0.739 \\
%          \textbf{XLM-RoBERTa} & 10 & 0.8 \\
%          \bottomrule
%     \end{tabular}
%     }
%     \caption{Transformer architectures scores}
%     \label{tab:transformers_comparison_small}
% \end{table}

The number of epochs was chosen according to the following criteria: the model does not overfit on the train data and shows  high results on the development data. To this end, we used early-stopping. For \texttt{ruBERT-tiny} even 50 epochs were not sufficient for reaching results  comparable to other models' performances.

\begin{table}[ht]
    \centering
    \resizebox{\linewidth}{!}{
    \begin{tabular}{lcccccc}
    \toprule
    \textbf{Models} & \textsc{char} & \textsc{fac} & \textsc{loc} & \textsc{org} & \textsc{per}  &  \textbf{Overall}  \\
    \midrule
    \texttt{ruBERT-tiny} & 0.712 & 0.8 & 0.748 & \textbf{0.4} & 0.738 & 0.731 \\
    \texttt{ruBERT} & 0.757 & \textbf{1.0} & \textbf{0.793} & \textbf{0.4} & \textbf{0.854} & \textbf{0.813} \\
    \texttt{ruRoBERTa} & 0.703 & 0.333 & 0.729 & 0.166 & 0.795 & 0.739 \\
    \texttt{XLM-RoBERTa} & \textbf{0.817} & 0.363 & 0.742 & 0.333 & 0.825 & 0.8\\
    \bottomrule
    \end{tabular}
    }
    \caption{Transformer architectures F1-scores}
    \label{tab:transformers_comparison_large}
\end{table}

% \dasha{\foreignlanguage{russian}{done - нужна конфьюжен матрица с доменами для трансформеров}}

According to Figure \ref{fig:image}, \textsc{char} and \textsc{per} entities were mostly wrongly detected as \textsc{o} by natasha, SpaCy and Stanza annotators. \textsc{org} tags were also erroneously detected by these parsers, which was quite similar to the results of transformer models' results. \textsc{loc} tags almost in all cases were detected correctly both by pre-trained parsers transformer models, while \textsc{fac} tags were significantly better found by the former ones.

According to Figure \ref{fig:confusion_transformers}, \texttt{XLM-RoBERTa}'s performance could be considered quite successful: \textsc{char} tags, as well as \textsc{per} and \textsc{loc}, were almost infallibly predicted. More exactly, \textsc{per} entity was never predicted as another entity on test data. \textsc{fac} entity was mixed with ORG tag in \texttt{XLM-RoBERTa}’s predictions while \textsc{org} tag itself is nearly in all cases is considered as \textsc{o}  tag by the model.

Figure \ref{fig:confusion_transformers} also presented \texttt{ruBERT-tiny}’s performance: \textsc{char} and \textsc{org} entities were  erroneously predicted as \textsc{o} more often, if compared to  \texttt{XLM-RoBERTa}. Nebertheless, in most cases the model predicts correctly. \texttt{ruBERT-tiny} extracted all \textsc{fac} and almost all \textsc{per} tags without major errors. 

As for \texttt{ruBERT}’s results, \textsc{o} tags were rarely misclassified as \textsc{char}, while all other tags were predicted entirely correctly or with inconsequential mistakes.

\texttt{ruRoBERTa}’s performance was far from being perfect, as \textsc{o}-entities were heavily confused with other tags, but most predictions of  other entities were correct.

As for major tendencies in models’ predictions, we can notice that \textsc{org} entity in most cases was detected as \textsc{o}  tag which although was not desired, but still can encourage us to re-analyze \textsc{org} entities and collect substantially more examples of \textsc{org} tag occurrence. \textsc{fac} entities were either (in most cases) correctly predicted, or mispredicted as \textsc{org}. \textsc{o}  tags were sometimes detected as \textsc{per} entity.

\section{Conclusion}

This paper introduces Razmecheno, a novel dataset for Named Entity Recognition. The texts in the dataset are sampled from the project ``Prozhito'', which comprises personal diaries, written in Russian, from the 17th century up to the end of the 20th century. In particular, texts, marked up in Razmecheno belong to the mid-1980 years, the period in Russia, commonly known as Perestroika. Razmecheno is a middle-scale dataset so that it contains enough data to carry out literal and historical studies. 

The annotation schema, used in Razmecheno, is simplistic. It consists of five named entity types, of which four are commonly used in NER datasets, namely, \textbf{persons}, \textbf{locations}, \textbf{organization}, and \textbf{facilities}. An only named entity type, introduced in this project, is \textbf{characteristics} of the different groups of people. The annotations are flat; overlapped, or nested entities are not allowed at the moment. 

As our annotation schema matches a commonly used inventory of named entity types, it is possible to leverage upon pre-trained models and transfer learning techniques.  The experimental evaluation of Razmecheno is two-fold. First, we carry out an extensive analysis of how available off-the-shelf NER tools cope with the task. The results reveal, that Natasha outperforms other tools under consideration by a small margin. However, of five named entity types, the off-the-shelf tools used to support only three. Next, we experiment with four state-of-the-art pre-trained Transformers. A monolingual model, \texttt{ruBERT} significantly outperforms other Transformers, followed by a multilingual model \texttt{XLM-RoBERTa}.    

There are a few directions for Razhmecheno development. We plan to annotate the collected sentences for other information extraction tasks, including co-reference resolution, relation extraction, and entity linking. This way, Razhmecheno could serve as a test-bed for end-to-end information extraction models. Experiments in domain adaptation and cross-lingual transfer from other languages are another research line. Finally, we have set up the whole environment to annotate texts from ``Prozhito'', so that diaries from other periods can be marked up with a little effort.

\nocite{*}
\section{Bibliographical References}
%\label{main:ref}
\bibliographystyle{lrec2022-bib}
\bibliography{bibliography}

\section{Language Resource References}
\label{lr:ref}
\bibliographystylelanguageresource{lrec2022-bib}
\bibliographylanguageresource{custom}

\onecolumn
\clearpage
\appendix

\section{Models performance on different datasets}
\label{app:comparison}
\begin{table}[h]
\centering
\resizebox{\textwidth}{!}{
\begin{tabular}{lllllllllllll}
\toprule
                \multirow{2}{*}{\textbf{Models}}  & \multicolumn{3}{c}{\textbf{factru}}                                                                                                            & \multicolumn{3}{c}{\textbf{ne5}}                                                                                                                                                                                           & \multicolumn{3}{c}{\textbf{bsnlp}}                                                                                                                                                                                         & \multicolumn{3}{c}{\textbf{razmecheno}}                                                                                                 \\ 
                 %\cmidrule{2-13}
                 & \multicolumn{1}{c}{\cellcolor[HTML]{FFFFFF}{\color[HTML]{24292F} \textsc{per}}} & \multicolumn{1}{c}{\cellcolor[HTML]{FFFFFF}{\color[HTML]{24292F} \textsc{loc}}} & \multicolumn{1}{c}{\cellcolor[HTML]{FFFFFF}{\color[HTML]{24292F} \textsc{org}}} & \multicolumn{1}{c}{\cellcolor[HTML]{FFFFFF}{\color[HTML]{24292F} \textsc{per}}} & \multicolumn{1}{c}{\cellcolor[HTML]{FFFFFF}{\color[HTML]{24292F} \textsc{loc}}} & \multicolumn{1}{c}{\cellcolor[HTML]{FFFFFF}{\color[HTML]{24292F} \textsc{org}}} & \multicolumn{1}{c}{\cellcolor[HTML]{FFFFFF}{\color[HTML]{24292F} \textsc{per}}} & \multicolumn{1}{c}{\cellcolor[HTML]{FFFFFF}{\color[HTML]{24292F} \textsc{loc}}} & \multicolumn{1}{c}{\cellcolor[HTML]{FFFFFF}{\color[HTML]{24292F} \textsc{org}}} & \multicolumn{1}{l}{\textsc{per}}           & \multicolumn{1}{l}{\textsc{loc}}                                                          & \textsc{org}           \\ \cmidrule(lr){2-4}  \cmidrule(lr){5-7}\cmidrule(lr){8-10}\cmidrule(lr){11-13}
\texttt{SpaCy}   & \multicolumn{1}{l}{0.901}                                              & \multicolumn{1}{l}{0.886}                                              & 0.765                                                               & \multicolumn{1}{l}{0.967}                                              & \multicolumn{1}{l}{0.928}                                              & 0.918                                              & \multicolumn{1}{l}{0.919}                                    & \multicolumn{1}{l}{0.823}                                              & 0.693                                             & \multicolumn{1}{l}{0.64}          
        & \multicolumn{1}{l}{0.54}                                                & 0.16 \\ 
        %\midrule
\texttt{Stanza}  & \multicolumn{1}{l}{0.943}                                              & \multicolumn{1}{l}{0.865}                                              & 0.687                                                                   &  \multicolumn{1}{l}{0.923}                                              & \multicolumn{1}{l}{0.753}                                              & 0.734                                             & \multicolumn{1}{l}{0.938}                                     & \multicolumn{1}{l}{0.838}             & 0.724                                  
    & \multicolumn{1}{l}{0.69}          & \multicolumn{1}{l}{0.4}       & 0.11          \\ %\midrule
\texttt{Natasha} & \multicolumn{1}{l}{{\color[HTML]{24292F} \textbf{0.959}}}              & \multicolumn{1}{l}{{\color[HTML]{24292F} \textbf{0.915}}}              & {\color[HTML]{24292F} \textbf{0.825}}                                  & \multicolumn{1}{l}{{\color[HTML]{24292F} \textbf{0.984}}}              & \multicolumn{1}{l}{{\color[HTML]{24292F} \textbf{0.973}}}              & {\color[HTML]{24292F} \textbf{0.951}}                                   & \multicolumn{1}{l}{{\color[HTML]{24292F} \textbf{0.944}}}              & \multicolumn{1}{l}{0.834}                                              & 0.718                                                                   & \multicolumn{1}{l}{0.77} & \multicolumn{1}{l}{\cellcolor[HTML]{FFFFFF}{\color[HTML]{212121} 0.54}} & 0.14          \\
%\midrule
\texttt{ruBERT-tiny}
& \multicolumn{1}{l}{0.619}              & \multicolumn{1}{l}{0.395}
& 0.558                                 & \multicolumn{1}{l}{0.619}          & \multicolumn{1}{l}{0.414}             & 0.564                                   & \multicolumn{1}{l}{0.318}             & \multicolumn{1}{l}{0.333}                                              & 0.180                                                                  & \multicolumn{1}{l}{0.738} & \multicolumn{1}{l}{0.748} & \textbf{0.4}          \\
%\midrule
\texttt{ruBERT}
& \multicolumn{1}{l}{0.548}             & \multicolumn{1}{l}{0.358}              & 0.461                                  & \multicolumn{1}{l}{0.883}           & \multicolumn{1}{l}{0.777}              & 0.856                                   & \multicolumn{1}{l}{0.483}             & \multicolumn{1}{l}{{0.451}}                                              & 0.423                                                                  & \multicolumn{1}{l}{\textbf{0.854}} & \multicolumn{1}{l}{\textbf{0.793}} & \textbf{0.4}          \\
%\midrule
\texttt{ruRoBERTa}
& \multicolumn{1}{l}{0.468}              & \multicolumn{1}{l}{0.261}            & 0.406                                  & \multicolumn{1}{l}{0.768}           & \multicolumn{1}{l}{0.593}              & 0.687                                   & \multicolumn{1}{l}{0.192}             & \multicolumn{1}{l}{{0}}                                              & 0                                                                   & \multicolumn{1}{l}{0.795} & \multicolumn{1}{l}{0.729} & 0.166          \\
%\midrule
\texttt{XLM-RoBERTa}
& \multicolumn{1}{l}{0.879}              & \multicolumn{1}{l}{0.763}              & 0.78                                  & \multicolumn{1}{l}{0.963}           & \multicolumn{1}{l}{0.936}              & 0.944                                   & \multicolumn{1}{l}{0.762}             & \multicolumn{1}{l}{\textbf{0.899}}                                              & \textbf{0.726}                                                                   & \multicolumn{1}{l}{0.825} & \multicolumn{1}{l}{0.742} & 0.333         \\
\bottomrule
\end{tabular}
}
\caption{See Section \ref{sec:nerus_review} for the review of these corpora in the Nerus suite. The data on the performance for off-the-shelf were taken from Natasha project
\protect\footnotemark}
\label{tab:offtheshelfperformance}
\end{table}
\footnotetext{\url{https://github.com/natasha/slovnet\#ner}}

\clearpage
\section{Off-the-Shelf models' span recognition}
\label{app:spans_analysis}
To evaluate how precise off-the-shelf models are in span recognition, we divide all cases of recognition in 11 groups:
\begin{itemize}
    \item \textbf{left more}: the right border of a span was detected correctly but on the left border a model included more words than in our annotation;
    \item \textbf{right more}: more words were included into a span on the right side;
    \item \textbf{left less}: the right border was correctly detected but on the left side one or more words were missing;
    \item \textbf{right less}: the left border was detected but on the right side less words were included;
    \item \textbf{more}: on both sides a model annotated more words than in the data;
    \item \textbf{less}: on the both sides a model detected a smaller span;
    \item \textbf{equal}: a model detected a span correctly;
    \item \textbf{left right}: the borders of a span were shifted from left to right, i.e. on the left side less words were included and on the right side a model detected some extra words;
    \item \textbf{right left}: the borders of a span were shifted from right to left;
    \item \textbf{not found}: models did not find a span or annotated it with a wrong tag;
    \item \textbf{false detected}: models found spans that were not in the manual annotation.
    
\end{itemize}

\begin{figure}[h]
\centering
\includegraphics{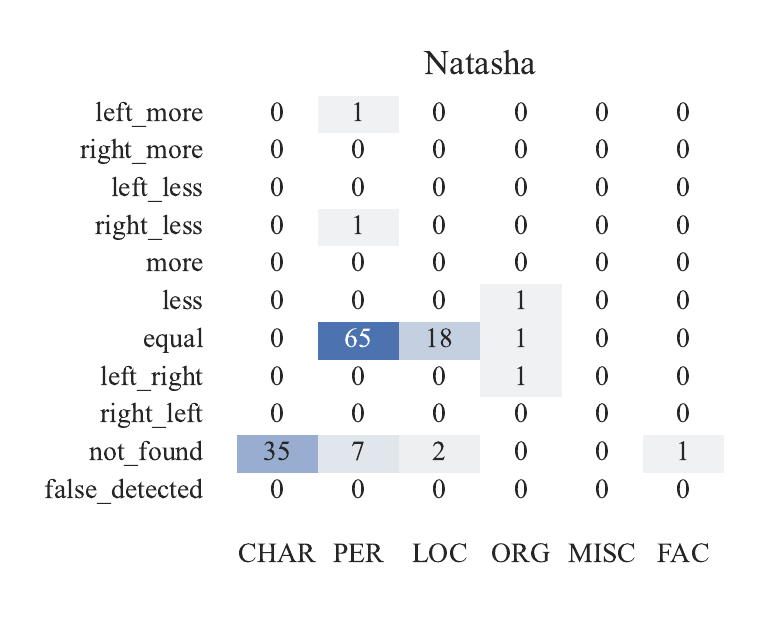}
\includegraphics{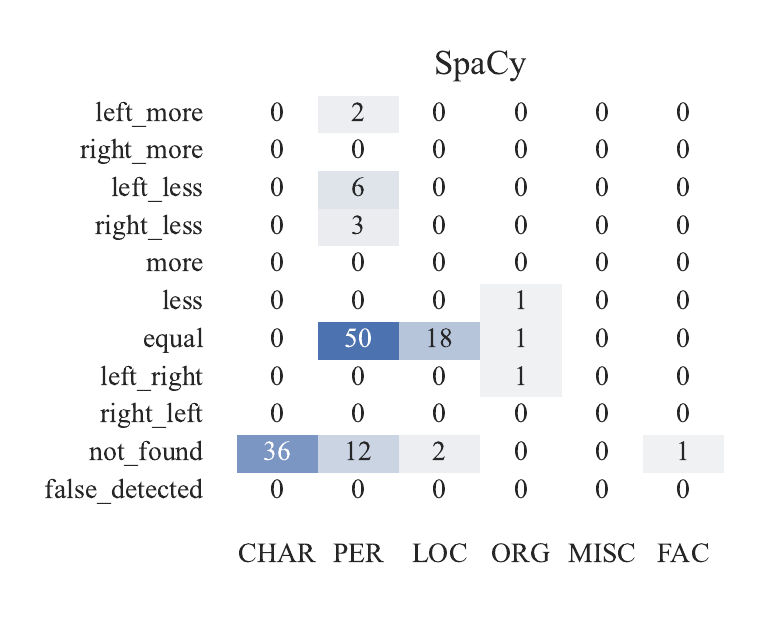}
\includegraphics{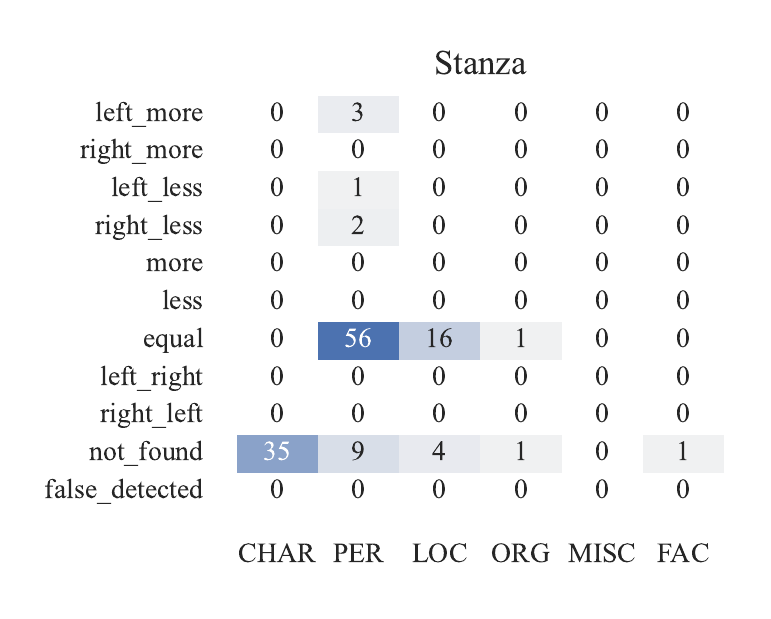}
\caption{Off-the-shelf tools' mistakes in span recognition for each entity}
\label{fig:spans_confusion}
\end{figure}

Figure \ref{fig:spans_confusion} shows the absolute number of cases of each type described above.

\clearpage
\section{Transformers hyper-parameters}
\label{app:trans_hyperparam}
\begin{table}[h]
    \centering
    % \resizebox{0.45\textwidth}{!}{
    \begin{tabular}{lccc}
    \toprule
    \textbf{Models} & \textbf{Number of epochs} & \textbf{Learning rate} & \textbf{Weight decay}  \\
    \midrule
    \texttt{ruBERT-tiny} & 50 & 1e-5 & 3e-5  \\
    % \midrule
    \texttt{ruBERT} & 10 & 1e-4 & 2e-5  \\
    % \midrule
    \texttt{ruRoBERTa} & 5 & 1e-5 & 2e-5 \\
    % \midrule
    \texttt{XLM-RoBERTa} & 10 & 3e-5 & 1e-4\\
    \bottomrule
    \end{tabular}
    % }
    \caption{Transformer architectures' hyperparameters}
    \label{tab:hyperparams}
\end{table}

% \clearpage
\section{Transformers confusion matrix}
\label{app:trans_conf_mat}

\begin{figure}[h]
\centering
\includegraphics{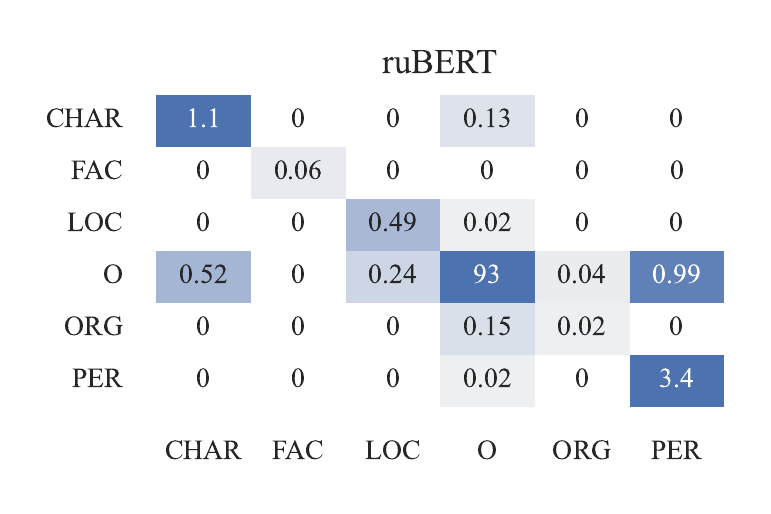}
\includegraphics{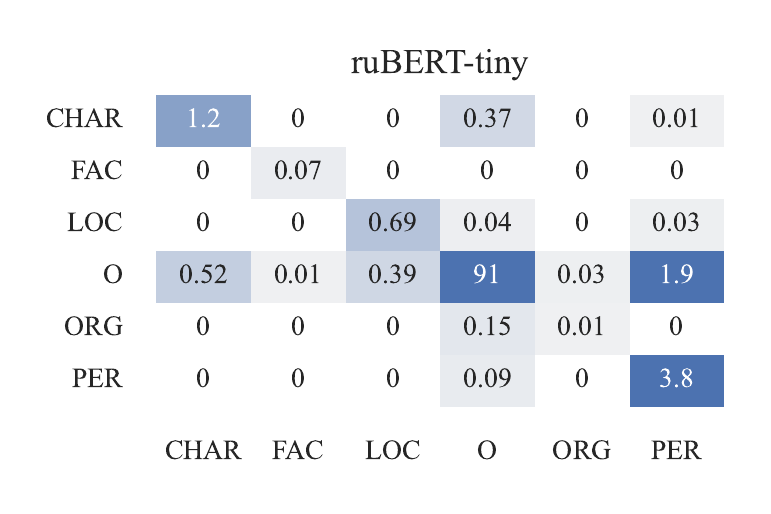}
\includegraphics{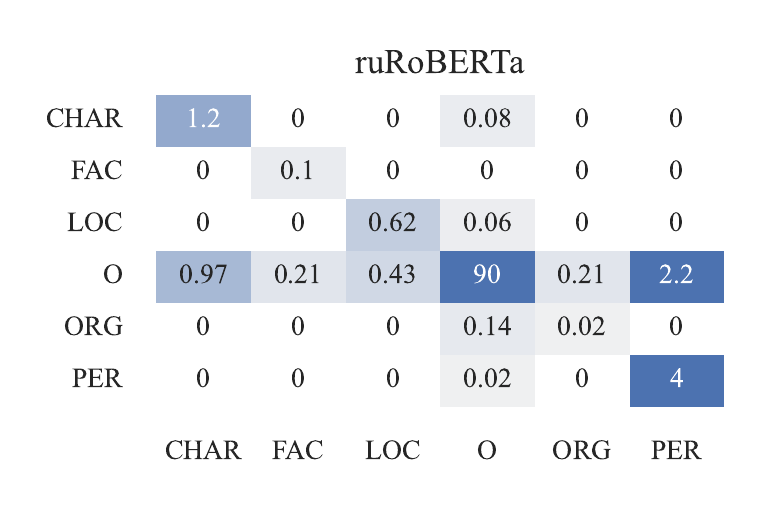}
\includegraphics{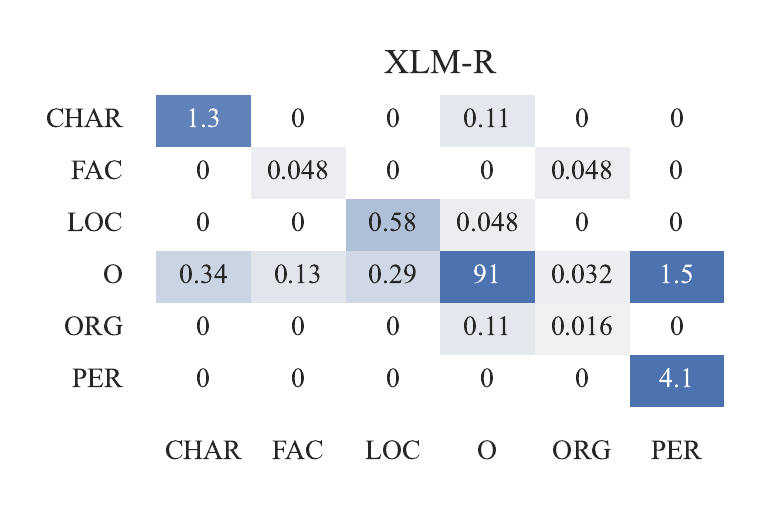}
\caption{Confusion matrix of ruBERT, ruBERT-tiny, ruRoBERTa and XLM-RoBERTa models' results on the test dataset}
\label{fig:confusion_transformers}
\end{figure}

\end{document}